%% file: main.tex
\documentclass[10pt,twocolumn,letterpaper]{article}

\usepackage{cvpr}              
\usepackage[accsupp]{axessibility}


\usepackage{amsmath}
\usepackage{subcaption}
\usepackage{multirow}

\usepackage{pifont}
\usepackage{amssymb}
\usepackage{makecell}
\usepackage{multicol}
\usepackage{marvosym}

\usepackage[ruled]{algorithm2e}             
\usepackage{array}

\newcommand{\xmark}{\ding{55}}%

\def\eg{\emph{e.g.,}}

\def\ie{\emph{i.e.,}}

\newcommand{\Ours}{\texttt{MovieBench}\xspace}

\definecolor{mygray}{gray}{.92}

\definecolor{cvprblue}{rgb}{0.21,0.49,0.74}
\usepackage[pagebackref,breaklinks,colorlinks,allcolors=cvprblue]{hyperref}

\def\paperID{\#4522} 
\def\confName{CVPR}
\def\confYear{2025}

\title{\Ours: A Hierarchical Movie Level Dataset for Long Video Generation}

\author{
{\large
Weijia Wu$^1$},
{\large
Mingyu Liu$^2$},
{\large
Zeyu Zhu$^1$},
{\large
Xi Xia$^1$},
{\large
Haoen Feng$^1$},
{\large
Wen Wang$^2$},
{\large
Kevin Qinghong Lin$^1$},\\
{\large
Chunhua Shen$^2$},
{\large
Mike Zheng Shou$^1$$^{({\textrm{\Letter}})}$}\\
\\
{\large
$^1$Show Lab, National University of Singapore$ \qquad $}\\
{\large
$^2$Zhejiang University $ \qquad $
}
}

\begin{document}
\maketitle
\let\thefootnote\relax\footnotetext{$^{\textrm{\Letter}}$ Corresponding author.}
\input{./0_abstract}

\input{./1_intro}
\input{./2_related_works}

\input{./3_dataset}
\input{./4_experiment}

\input{./5_conclusion}

\section*{Acknowledgements} 

This research is supported by the Ministry of Education,
Singapore, under the Academic Research Fund Tier 1
(FY2023). 
M. Liu, C. Shen's participation was supported by the National Key R\&D Program of China (No.\  2022ZD0160101). 
\input{./X_suppl}

{
    \small
    \bibliographystyle{ieeenat_fullname}
    \bibliography{main}
}


\end{document}

%% file: 0_abstract.tex
\begin{abstract}
Recent advancements in video generation models, like Stable Video Diffusion, show promising results, but primarily focus on short, single-scene videos.
These models struggle with generating long videos that involve multiple scenes, coherent narratives, and consistent characters. 
Furthermore, there is no publicly available dataset tailored for the analysis, evaluation, and training of long video generation models.
In this paper, we present \Ours: A Hierarchical Movie-Level Dataset for Long Video Generation, 
which addresses these challenges by providing unique contributions: 
(1) movie-length videos featuring rich, coherent storylines and multi-scene narratives, 
(2) consistency of character appearance and audio across scenes, 
and (3) hierarchical data structure contains high-level movie information and detailed shot-level descriptions.
%
%
Experiments demonstrate that \Ours brings some new insights and challenges, such as maintaining character ID consistency across multiple scenes for various characters. 
The dataset will be public and continuously maintained, aiming to advance the field of long video generation.
Data can be found at:
\href{https://weijiawu.github.io/MovieBench/}{\color{blue}{\tt MovieBench}}.

%
%
%

\end{abstract}

%% file: 1_intro.tex
\begin{figure}[t]
    \includegraphics[width=0.99\linewidth]{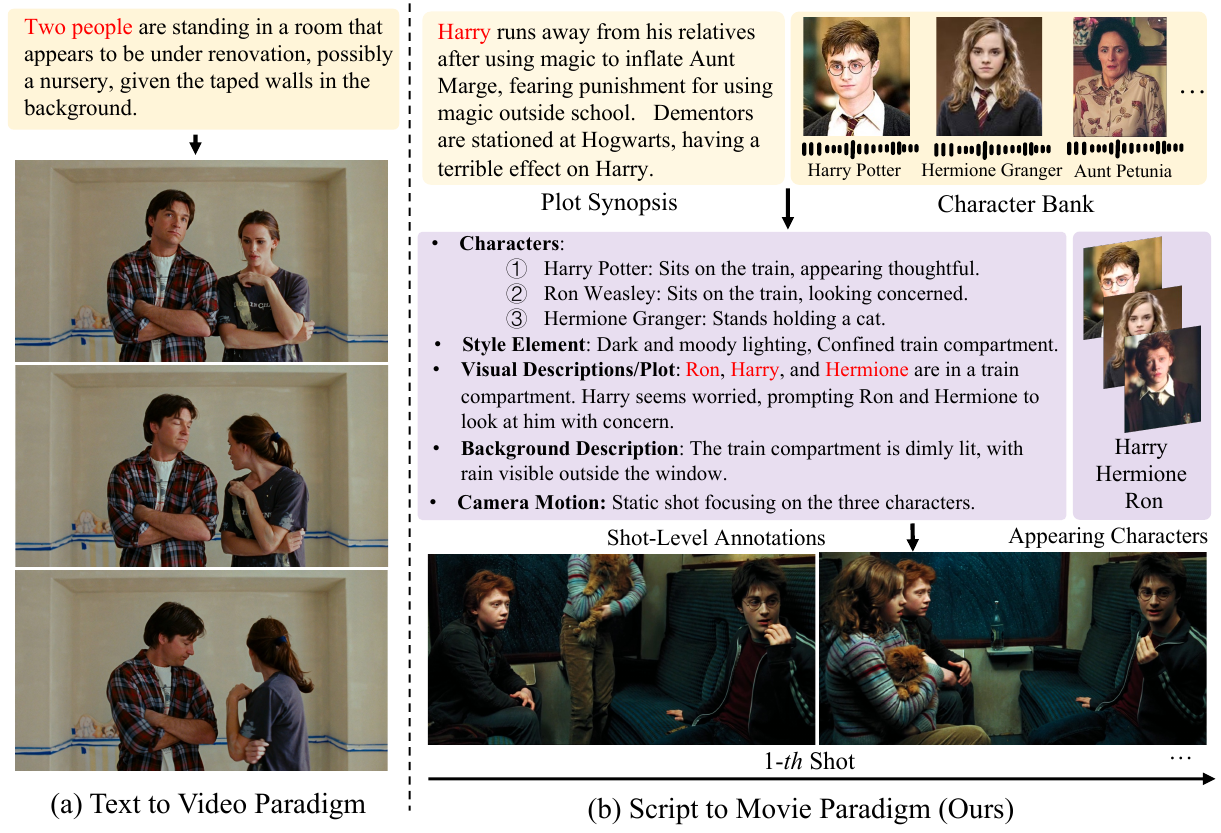}
	\vspace{-0.2cm}
	\caption{\textbf{Video Generation $v.s$ Movie Generation.} 
     The text-to-video paradigm (MiraData~\cite{ju2024miradata}) takes a text input without character information and generates a short video.
     In contrast, script-to-movie generation involves a complex storyline, requiring character consistency, plot progression, and audio synchronization.
    }
    \vspace{-0.2cm}
\label{motivation}
\end{figure}

\section{Introduction}
Video generation has seen rapid advancements in recent years, driven by improvements in generative models~\cite{videoworldsimulators2024,svd,esser2023structure,polyak2024movie}, data scale~\cite{ju2024miradata,internvid,nan2024openvid} and computational power.
Early successes in this domain were primarily based on diffusion process, such as Stable Video Diffusion~\cite{svd}, Video LDM~\cite{blattmann2023align} and I2vgen-xl~\cite{zhang2023i2vgen},  have demonstrated impressive results in generating high-quality short videos.
Recently, spatial-temporal transformer models, exemplified by Sora~\cite{videoworldsimulators2024}, have 
shown stronger performance by capturing both spatial and temporal dependencies within video sequences.
However, these approaches have mostly been applied to short videos (text-video paradigms), typically limited to single scenes without intricate storylines or character development, as shown in Figure~\ref{motivation} (a).

Despite recent advancements, generating long videos that maintain \textbf{character consistency}, cover \textbf{multiple scenes}, and follow a \textbf{rich narrative} remains an unsolved problem. 
Existing models fail to address the challenges of long video generation, including the need for maintaining character identity and ensuring logical progression through multiple scenes.
Moreover, a major bottleneck lies in the limitations of current benchmarks, which still focus on the analysis, training, and evaluation of short videos. 
Datasets like WebVid-10M~\cite{webvid10m}, Panda-70M~\cite{panda70m}, and HD-VILA-100M~\cite{hdvila} primarily consist of short video clips ranging from $5$ to $18$ seconds.
While recent efforts like MiraBench~\cite{ju2024miradata} have begun exploring longer video generation, the majority of the videos provided are still under one and a half minutes in length.
More importantly, these benchmarks lack the crucial annotations required for long video generation, such as character ID information and the contextual relationships between video clips.
The lack of character consistency and logical scene progression limits the development of models for long-form narrative generation.
As a result, the field is hindered by the absence of appropriate datasets tailored for movie/long video generation task.

To address these gaps, we introduce the \Ours dataset, specifically designed for movie-level long video generation (script-movie generation paradigms), as shown in Figure~\ref{motivation} (b).
\Ours provides three hierarchical levels of annotations: movie-level, scene-level, and shot-level.
At the \textbf{movie level}, the annotations focus on high-level narrative structures, such as script synopsis and a comprehensive character bank.
The character bank includes the name, portrait images, and audio samples of each character,
which can support tasks like custom audio generation, ensuring character appearance and audio consistency across multiple scenes.
\textbf{Scene-level} annotations encapsulate the progression of the story by detailing all the shots and events within a particular scene.
Scene categories help maintain consistency in background, foreground, style, and character outfits across multi-view videos within the same scene.
Finally, \textbf{shot-level} annotations capture specific moments, typically focusing on short sequences like close-ups or camera movements, usually lasting less than $10$ seconds.
Shot-level annotations typically include the characters present, plot, camera motion, background descriptions, and time-aligned subtitles and audio information for each video segment.
These annotations emphasize specific details of generated video, including the characters involved, their locations, dialogues, and actions, ensuring accurate alignment of this information.
\Ours consists of $91$ movies, with an average movie duration of $45.6$ minutes. 
\Ours seeks to advance research in long video generation, illing a major gap in current benchmarks.

To summarize, the contributions of this paper are: 
\begin{itemize}[leftmargin=*]
    \item We introduce \Ours, the first benchmark designed for movie-level video generation. 
    It establishes a new paradigm for creating coherent narratives and enabling multi-scene progression.

    \item \Ours provides three annotation levels: movie-level for scripts and character banks, scene-level for shot sequences and narratives, and shot-level for details like close-ups, plot, and camera movements.
    
    \item \Ours provides character consistency and coherent narrative development.
    Character banks include profiles with headshots, names, and audio samples, along with shot-level character sets.

    
    \item Experiments demonstrate that \Ours brings some new insights and challenges, such as maintaining character ID consistency, multi-view character ID coherence, and synchronized video generation with audio. 
\end{itemize}

%% file: 2_related_works.tex
\begin{table*}[t]
    \centering
    \small 
    \setlength{\tabcolsep}{1mm}
    \caption{\textbf{Comparison of \Ours and previous datasets.} `w', `m', `s', and `hr' refer to the $\text{words}$, $\text{minutes}$, $\text{seconds}$, and $\text{hours}$, respectively. 
    \Ours offers a unique contribution with character consistency, coherent storylines, and a hierarchical data structure.
    }
    \input{StatisticalData}
    \label{tab:query}
\end{table*}

\section{Related Works}

\subsection{Video-Text Datasets}
Numerous video-text datasets have been developed, initially focusing primarily on video understanding~\cite{han2023autoad,huang2020movienet,youcook2,vatex,wu2024end}, where research progress has been ahead of the video generation~\cite{ju2024miradata,panda70m}.
MSR-VTT~\cite{msrvtt}, TextVR~\cite{wu2025large}, ActivityNet~\cite{activitynet}, 
BOVText~\cite{ENCHMARKS2021_b6d767d2}, How2~\cite{how2}, and VALUE~\cite{li1value} paved the way for advancing tasks like video retrieval, captioning, video text spotting, and video question answering. 
These datasets, while extensive in their scope and impact, are typically framed around short-form video content with an emphasis on understanding rather than generation.

More recent works, such as MAD~\cite{soldan2022mad}, AutoAD~\cite{han2023autoad}, AutoAD II~\cite{han2023autoad2}, and AutoAD III~\cite{han2024autoad}, focus on movie-level understanding and descriptions, capture the complexity of long-form video content. 
These datasets enhance the understanding of movies by offering comprehensive narrative, character, and scene descriptions. 
However, while they excel in video retrieval and description tasks, their focus remains largely on understanding rather than generation, where significant challenges still exist, for example, how to generate multiple scenes, coherent narratives, and consistent characters.

\subsection{Video Generation}
The field of video generation has experienced rapid advancements, both in terms of models~\cite{videodiffusionmodel,hong2022cogvideo,kondratyuk2023videopoet,wu2025draganything,zhao2025motiondirector} and datasets~\cite{panda70m,webvid10m}, particularly in the generation of short video clips from textual descriptions.
On the model front, numerous outstanding works have emerged, such as diffusion-based models like SVD~\cite{svd}, VDM~\cite{videodiffusionmodel}, and SORA~\cite{videoworldsimulators2024}, as well as autoregressive models like VideoGPT~\cite{yan2021videogpt}, CogVideo~\cite{hong2022cogvideo}, and VideoPoet~\cite{videopoet}.
%
%
On the data front, notable datasets like WebVid-10M~\cite{webvid10m}, Panda-70M~\cite{panda70m}, HD-VILA-100M~\cite{howto100m}, and InternVid~\cite{internvid} have been instrumental in establishing large-scale video-text datasets.
WebVid-10M~\cite{webvid10m} has provided a substantial contribution to video generation tasks by offering a rich dataset of video-text pairs, enabling models to generate short, descriptive video clips. 
Similarly, Panda-70M~\cite{panda70m} and HD-VILA-100M~\cite{howto100m} expand on these efforts by incorporating diverse datasets, high-quality videos and more complex textual descriptions for generating visually and semantically rich video segments. 
MiraData~\cite{ju2024miradata}, meanwhile, focuses on enabling fine-grained video understanding and generation, incorporating dense annotations for improving both accuracy and diversity in generated clips.
%

However, these works are primarily designed for generating short videos, which do not meet the requirements for movie-level generation.
Movie-level generation is more complex, requiring the creation of longer video sequences while maintaining a coherent storyline, character consistency, and audio continuity.
Our \Ours try to address these challenges by introducing a hierarchical dataset specifically designed for movie-level generation. 
%


%% file: StatisticalData.tex
\begin{tabular}{l|c|cc|ccc|cccc}
    \multirow{2}{*}{Dataset} & \multirow{2}{*}{Subtitle} & \multicolumn{2}{c|}{Character}   & \multicolumn{3}{c|}{Avg Text Len / Avg Video Len}   & \multirow{2}{*}{Total video len} &  \multirow{2}{*}{Text}  & \multirow{2}{*}{Res.}  
    
    \\
    \cline{3-7}
    
     &  & Portrait & Audio   & Movie-Level & Scene-Level& Shot-Level   &  &    & 
     \\
    \hline
    
    HD-VG-130M~\cite{videofactory}    & \xmark & \xmark & \xmark & - & - & $\sim$9.6w / $\sim$5.1s    & $\sim$184Khr   &  Generated          & 720p     \\
    
    WebVid-10M~\cite{webvid10m}    & \xmark & \xmark & \xmark &  - & - & 12.0w / 18.0s   & 52Kh    & Alt-Text      & 360p     \\

    YouCook2~\cite{youcook2}  & $\checkmark$ & \xmark &  \xmark &  - & - & 8.8w / 19.6s & 176h  & Manual          & -     \\
    MSR-VTT~\cite{msrvtt}    & \xmark & \xmark &\xmark & -  & - & 9.3w / 15.0s & 40h  & Manual           & 240p  \\

    VATEX~\cite{vatex} & \xmark &    \xmark &\xmark &-  & - & 15.2w / $\sim$10.0s    & $\sim$115hr   & Manual      & -  \\

    
    Panda-70M~\cite{panda70m}  & \xmark & \xmark &\xmark & - & - & 13.2w / 8.5s  & 167Khr    & Generated      & 720p \\
    
    HD-VILA-100M~\cite{hdvila}  & \xmark & \xmark &\xmark &-  & - & 17.6w / 11.7s    & 760.3Khr   & ASR      & 720p  \\

    InternVid~\cite{internvid} & \xmark &\xmark &\xmark & - & - & 32.5w / 13.4s  & 371.5Khr   &  Generated                  & 720p  \\
    
    MiraData~\cite{ju2024miradata}   & \xmark & \xmark &\xmark & - & - & 318.0w / 72.1s    & 
    16Khr   &  Generated     & 1080p \\
    
    \hline
    \Ours ~(Ours)   & $\checkmark$ & $\checkmark$ & $\checkmark$ & 43.4Kw / 45.6m & 263.6w / 15.4s &  66.3w / 4.09s   & 
    69.2hr   &  Generated & 1080p \\

    \Ours ~(Ours++)   & $\checkmark$ & $\checkmark$ & $\checkmark$ & 51.5Kw / 43.8m & - &  79.5w / 4.01s   & 
    116.8hr   &  Generated & 1080p \\
    
    \end{tabular}

%% file: 3_dataset.tex
\section{\Ours Dataset}
\label{Method}

%

\begin{figure*}[t]
	\includegraphics[width=0.99\linewidth]{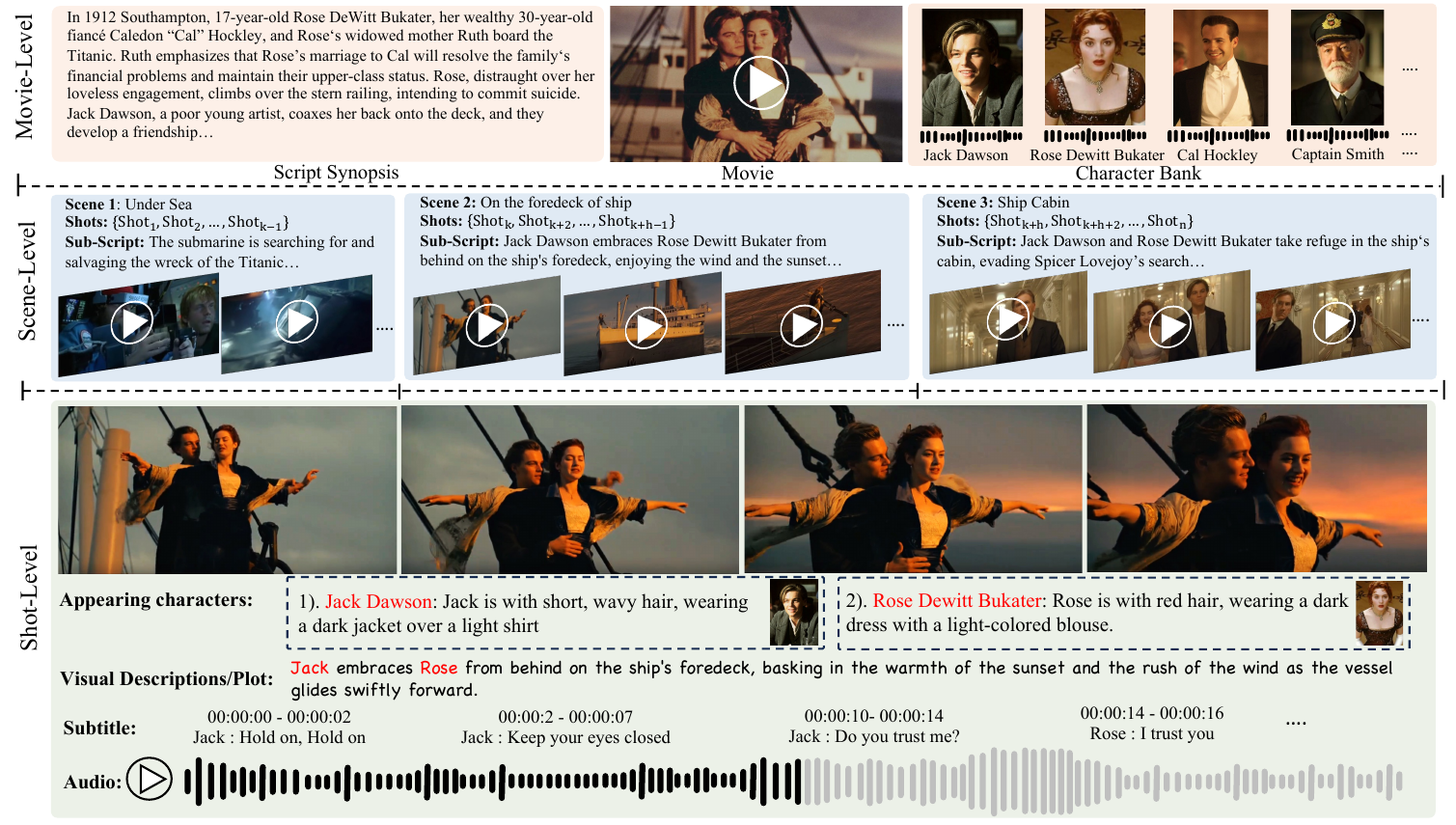}
	\vspace{-0.2cm}
	\caption{\textbf{\Ours Dataset.} 
     \Ours categorizes the movie annotations into three hierarchical data levels, representing different granularities of information: 
     1) Movie level provides a broad overview of the film; %
     2) Scene level provides mid-level scene consistency information; 
     3) Shot level emphasizes specific moments with detailed descriptions.
    }
    \vspace{-0.1cm}
\label{comparison1}
\end{figure*}

\begin{table}[t]
    \centering
    \small  \renewcommand\arraystretch{1.0}
    \setlength{\tabcolsep}{1mm}
    \caption{\textbf{Video Quality Comparison for Different Datasets.}}
    \vspace{-2mm}
    \input{video_quality}
    \label{tab:video_quality}
    \vspace{-2mm}
\end{table}


\subsection{Data Collection}
\label{datacollection}
For movie source of \Ours, we utilized $91$ movies from LSMDC~\cite{lsmdc}, which includes notable films such as `Harry Potter and the Prisoner of Azkaban'.
Using movies from LSMDC provides two key advantages:
1) Pre-existing Manual Annotations: 
MAD~\cite{soldan2022mad} provides manually annotated movie audio descriptions for the movie clip of LSMDC. 
These annotations can serve as valuable references to further enhance the accuracy of the generated annotations, as shown in Figure~\ref{comparison2}.
Note: Movie audio descriptions cannot be directly used as plot annotations due to lack of character consistency, narrative coherence.
2) Open-Source Video Data: 
The video data in LSMDC is publicly accessible, which allows us to avoid copyright risks, ensuring the long-term availability.
%
%
Due to limited human annotation resources, we currently collected a total of $91$ movies, with $85$ movies as the training set, and $6$ as the test set.
Table~\ref{tab:video_quality} demonstrates that \Ours offers advantages in both video quality and aesthetics.


\subsection{Movie Level Elements}
\label{MovieLevel}

\subsubsection{Script Synopsis}
The Script Synopsis plays a crucial role in offering a quick understanding of the high-level narrative structure of a movie.
%
%
For each movie in \Ours, we collect the corresponding Script Synopsis from publicly available source,\ie{} IMDb~\footnote{https://www.imdb.com/}.
On average, each synopsis contains approximately $1,542$ words, capturing the core elements of the film while offering sufficient detail to guide video generation tasks.
Script synopses can be used to generate scene and shot-level annotations with LLMs~(\eg{} GPT4-o), enhancing the efficiency of script-to-movie generation.

\subsubsection{Character Bank}
\label{bank}


Movie generation needs to maintain character consistency throughout the entire movie.
For the same character, when generating different scenes and shots, we usually require their face id to remain unchanged. 
Therefore, we are attempting to build a character bank.
We scraped the cast list from IMDB for each movie to verify the characters and their corresponding actors in the character bank.
Given a long-form movie $\mathcal{V}$, the character bank for this movie can be written as $\mathcal{B}_{\mathcal{V}} = \{[\texttt{char}_i, \texttt{act}_i, \mathcal{I}_i, \mathcal{A}_i]\}_{i=1}^{N}$,
where $N$ denotes the number of characters,
$\texttt{char}_i$ and $\texttt{act}_i$ are the $i$-th character and actor names in the movie, respectively.
$\mathcal{I}_i$ and $\mathcal{A}_i$ denotes the portrait images and audio samples of the character in the movie.

\begin{figure}[t]
	\includegraphics[width=0.99\linewidth]{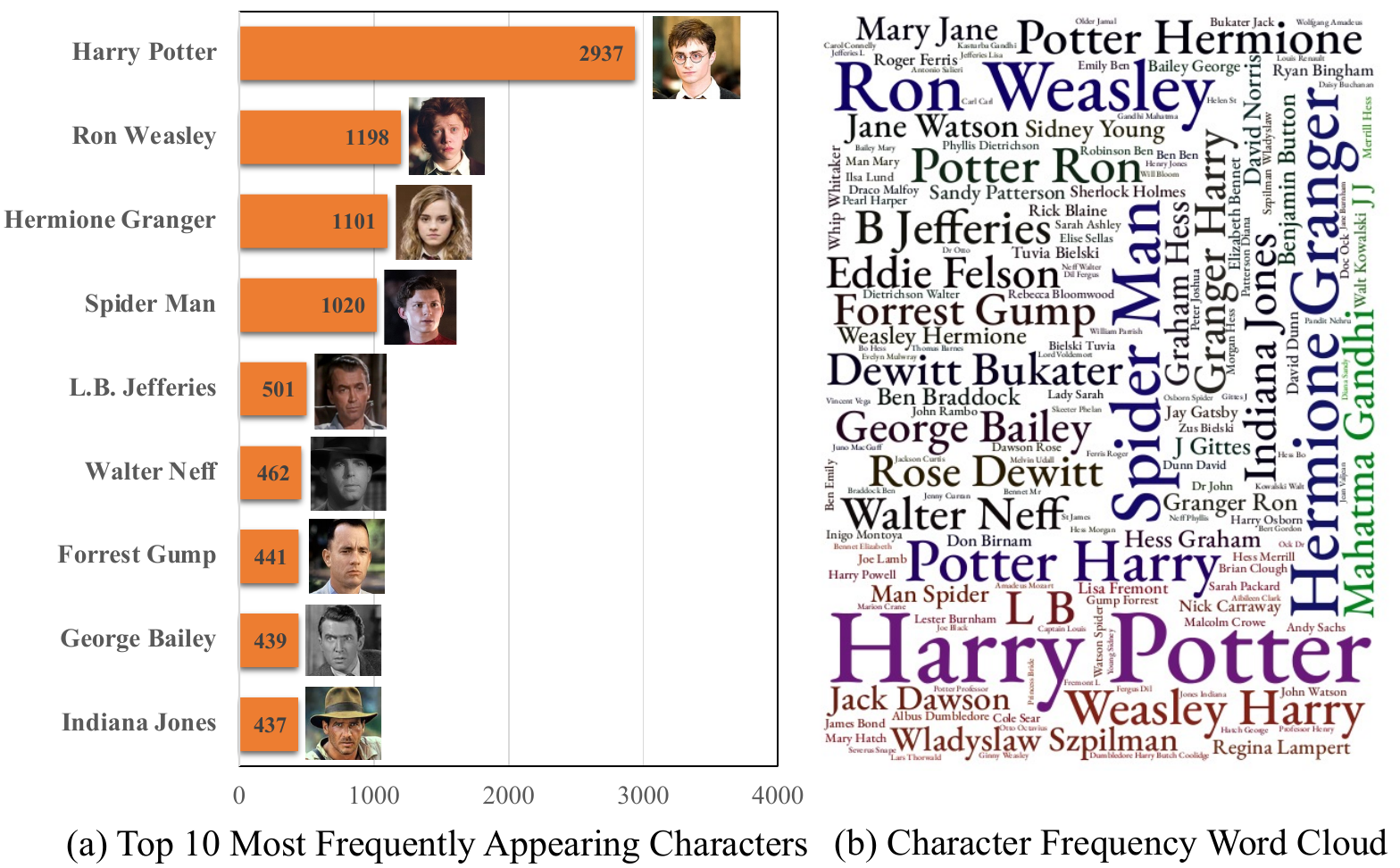}
	\vspace{-0.2cm}
	\caption{\textbf{Character Frequency Statistics.} 
     The frequency of different characters varies significantly.
    }
    \vspace{-0.05cm}
\label{CharacterStatistics}
\end{figure}

\textbf{Portrait Images of Characters}. With the character name, and stills of each character from the IMDB,
%
we used a object detector~(\ie{} GroundingDINO~\cite{liu2023grounding}) to detect all individuals in each still image and isolate the characters, ensuring that each picture contains only a single person.
For each character, we needed to select the corresponding still images and remove those not depicting the intended character.
Therefore, two annotators with a background in computer vision are invited to filter out images that were either incorrect or of low quality.
After the stills selection process, we also invited a professional quality inspector to conduct a quality check on the selected photos. 
Any non-compliant photos will be returned for re-annotation.
Figure~\ref{CharacterStatistics} shows the frequency of character appearances.
%
    
    
After completing the data cleaning process, a quality inspector was invited to conduct a quality check. 
$10$ randomly sampled movies were evaluated on two key aspects: portrait quality and portrait-name relevance. 
The inspector rated the quality of each portrait on a scale of $1$ to $5$, with $5$ as the highest.
The average portrait quality score was $4.53$, and portrait-name relevance scored $4.92$.
This demonstrates the high quality of the collected character bank. 
Detailed results can be found in the supplementary materials.


\textbf{Audio Samples of Characters}.
To collect audio samples for each character, we developed a structured process:
1) Face Detection and Recognition: a face detector~\cite{zhang2017s3fd} and recogniter~\cite{serengil2020lightface} are used to detect and recognize all faces by matching with $\{[\texttt{char}_i, \texttt{act}_i, \mathcal{I}_i]\}_{i=1}^{N}$ in each frame. 
%
%
2) Speaker Detection and Duration Identification: next, a speaker detector~\cite{liao2023light} was employed to determine which character is speaking, along with the duration.
3) Audio Extraction: based on the identified durations, we extracted audio segments corresponding to each character.
4) Quality Verification: finally, a annotator reviewed audios of each character to confirm that it indeed belongs to the intended character.
Any mismatched audio segments~(The predicted 057
speaker does not match the audio) were removed.
Ultimately, for each character, a noise-free audio sample was collected.
The annotation can be used in tasks such as audio customization and audio-driven video generation.


\subsection{Scene Breakdown}
\label{SceneLevel}
\begin{figure}[t]
	\includegraphics[width=0.99\linewidth]{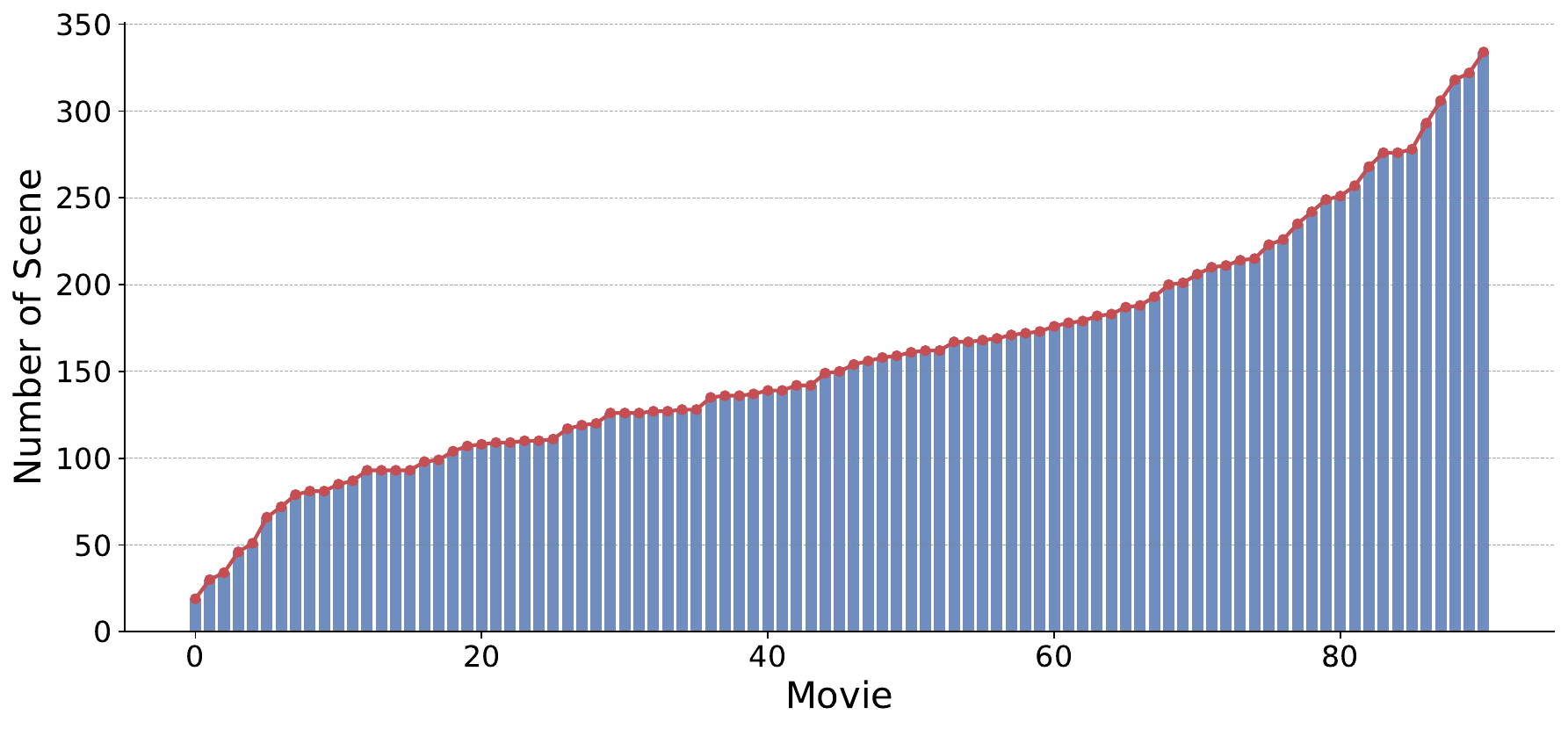}
	\vspace{-0.2cm}
	\caption{\textbf{Distribution of Scenes.} 
    The number of scenes varies significantly across different movies, from $20$ to $350$ scenes.
    }
    \vspace{-0.1cm}
\label{ScenesDistribution}
\end{figure}

The LSMDC movies are pre-segmented at the shot level, enabling us to classify each shot for scene breakdown.
Given $M$ shot-level video clips $\{S^0,S^1,\dots,S^m\}$, VLM~(\eg{} GPT-4-o) is used to obtain the corresponding scene breakdown results $\{R^0,R^1,\dots,R^m\}$.
Since adjacent video clips are likely to be the same scene, we began by classifying the first video clip in sequence. 
Then, we progressively classified the subsequent clips in chronological order.
For non-initial clips, the previous clip and its scene label as additional input, allowing the model to determine if the current clip belongs to the same scene.
If the current clip was not classified under the same scene, a new scene label was generated accordingly.
For $t$-th video clip $S^t$, the scene breakdown results $R^t$ can be formulated as:
\begin{equation}
\begin{split} 
R^t = \left\{
\begin{array}{lcl} \text{VLM}(S^t) & & t = 1 \\
\text{VLM}(S^t, S^{t-1}, R^{t-1}) & & t \geq 2 \end{array}
\right. 
\end{split}
\label{scene_equ}
\end{equation}
This iterative approach accurately identifies scene boundaries while ensuring consistency across related shots.
Figure~\ref{ScenesDistribution} shows the distribution of scene counts across different movies, ranging from $20$ to $350$ scenes.

\begin{figure}[t]
	\includegraphics[width=0.99\linewidth]{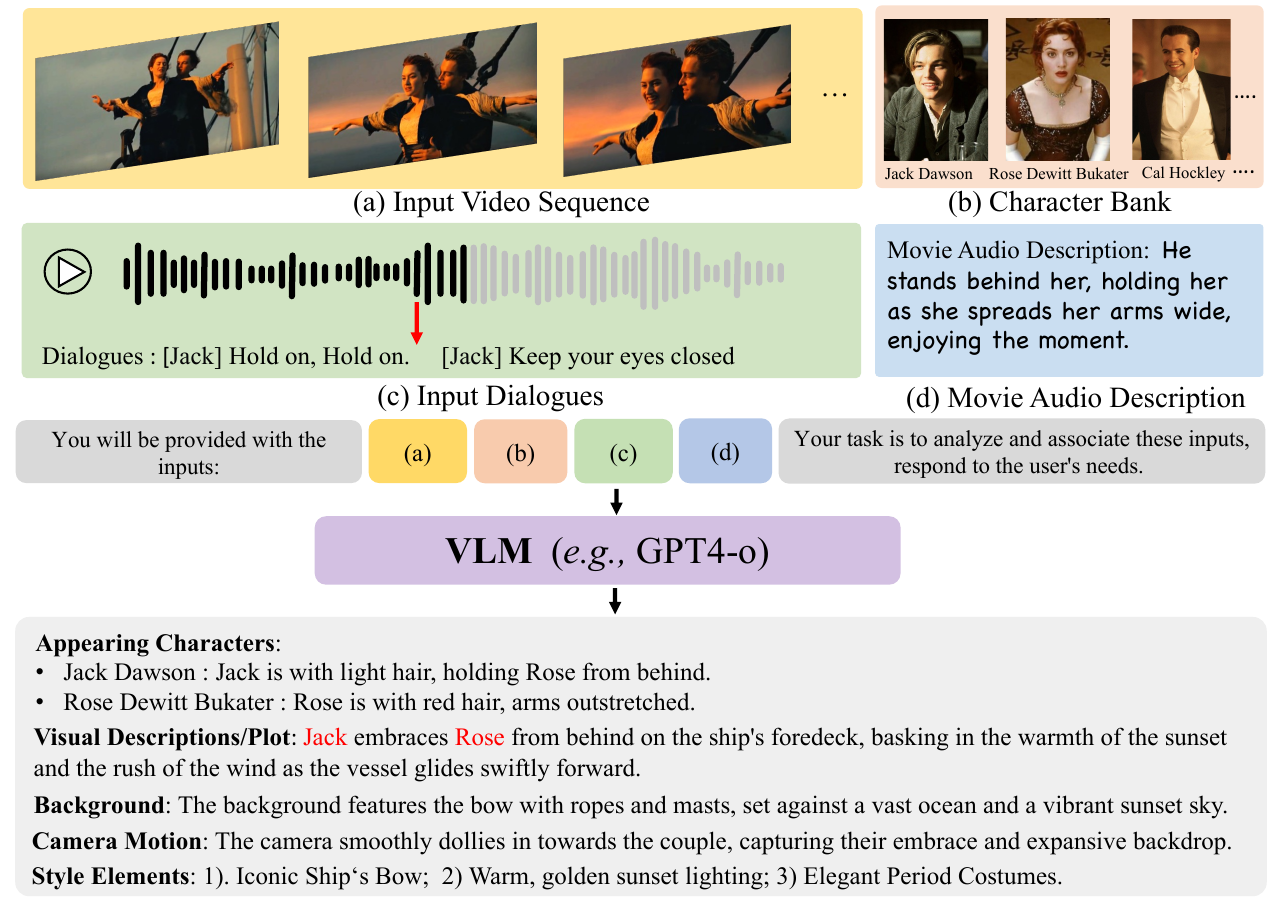}
	\vspace{-0.2cm}
	\caption{\textbf{Annotation Generation for Shot-Level Video.} 
     With video, character banks, audio, and movie descriptions, VLM can summarize the content, including characters and plot.
    }
\label{comparison2}
\end{figure}

\begin{table}[t]
    \centering
    \small 
    \setlength{\tabcolsep}{1mm}
    \caption{\textbf{Quality Evaluation for Shot-level Annotation.} }
    \vspace{-2mm}
    \input{EvalforShotLevel}
    \label{tab:EvalforShotLevel}
    \vspace{-2mm}
\end{table}

\subsection{Shot-Level Temporal Annotations}
\label{ShotLevel}

%

\subsubsection{Annotation Generation}
Based on the movie-related works~\cite{huang2020movienet,azzarelli2024reviewing}, when aiming to generate a shot-level video, certain annotations seems indispensable: 
1) Appearing Characters;
2) Plot;
3) Scene/background;
4) Shooting Style;
5) Camera Motion.
Some previous works have validated that vision-language models (VLMs)~\cite{chen2023pixart,wu2023paragraph,ju2024miradata} can generate accurate descriptive annotations.
Inspired by MovieSeq~\cite{lin2024learning}, VLMs (\eg{} GPT-4-o) was used to generate relevant annotations.
As shown in Figure~\ref{comparison2}, 
to enable the VLM to better understand and summarize the video clip, four types of information are provided as the interleaved multimodal input: 
1) Video frame includes visual information; 
2) Character Bank includes images and names of characters (Sec.~\ref{bank}).
3) Subtitles.
4) Movie Audio Description contains manually crafted visual descriptions.
Using the interleaved multimodal sequence described above, we generate a corresponding interleaved instruction and input it into VLM.
The detailed prompt strategy and interleaved sequence can be found in the supplementary materials.

\begin{figure*}[t]
	\includegraphics[width=0.99\linewidth]{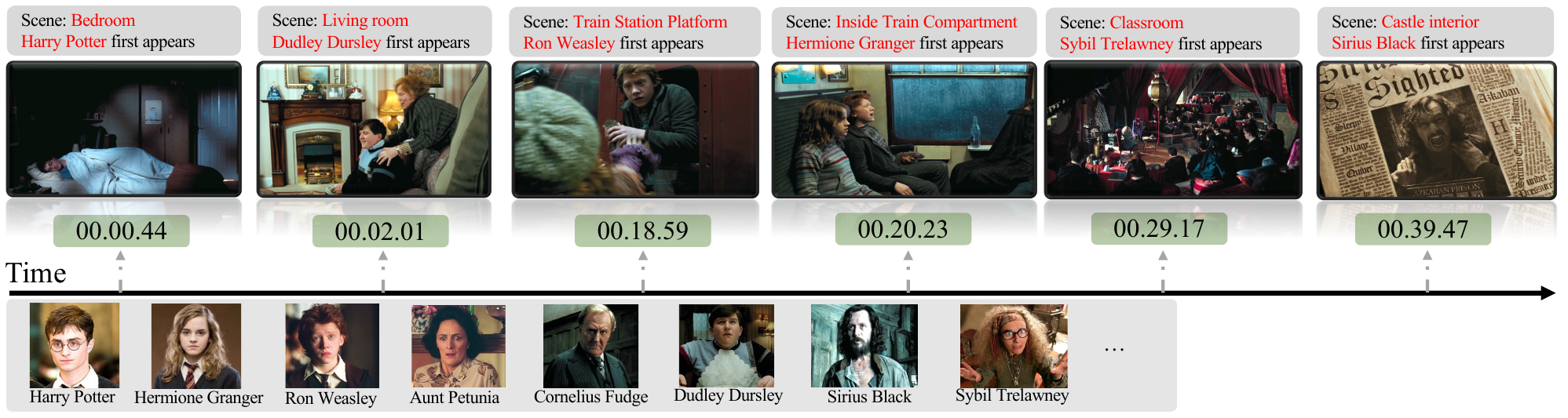}
	\vspace{-0.2cm}
	\caption{\textbf{Visualization of Scene Transitions and Character Appearance Order.} 
    \Ours is the first dataset to offer unique coherence and narrative progression across character relationships, scene transitions, and storyline development.
    }
\label{scene_flow1}
\end{figure*}


\subsubsection{Quality Evaluation and Correction}
The generated annotations~(\eg{} plot description) are not always accurate and may suffer from hallucinations~\cite{bai2024hallucination} or incomplete descriptions.
Accordingly, we randomly sampled $500$ shot-level video clips and the annotations for quality evaluation.
As shown in Table~\ref{tab:EvalforShotLevel}, we evaluated the quality of description-based annotations in terms of Completeness and Hallucination.
%
For appearing characters, we evaluate the performance using the character id consistency, \ie{} ${\rm F1\,Score_{ID}}$ (details see \S\ref{ShotLevel}), as shown in Table~\ref{tab:CharacterShotLevel}.

To further improve the accuracy of the generated annotations, we also enlisted two annotators with backgrounds in computer vision to refine the annotations for the test set. 
The details for correction and refinement guidelines can be found in the supplementary materials.
Due to limited annotator resources and the relatively high accuracy of the GPT-4-o generated annotations, we only performed manual corrections on the test set ($6$ movies). 
%

\begin{table}[t]
    \centering
    \small 
    \setlength{\tabcolsep}{1mm}
    \caption{\textbf{Quality Evaluation for Shot level Characters Set.}}
    \vspace{-2mm}
    \input{CharacterShotLevel}
    \label{tab:CharacterShotLevel}
    \vspace{-2mm}
\end{table}

\subsubsection{Audio and Subtitles}
To obtain character-specific subtitles and audio in alignment with video timing, a approach is implemented as follows:
1) Speaker Diarization and Segmentation: a speaker diarization tool~\cite{Plaquet23} is utilized to divide the continuous audio into distinct audio segments, each representing a independent speaker.
2) Matching via Audio Embeddings: For each segmented audio clip, the audio embedding was extracted with Pyannote~\cite{Plaquet23}.
Then cosine similarity matching is used to identify the character, by comparing the embedding with embeddings from the Character Audio Bank $\{[\texttt{char}_i, \texttt{act}_i, \mathcal{A}_i]\}_{i=1}^{N}$.
3) Subtitle Generation: Once character-specific audio segments were identified, Whisper~\cite{radford2023robust} can be used to transcribe the speech into subtitles.
%


\begin{figure*}[t]
	\includegraphics[width=0.99\linewidth]{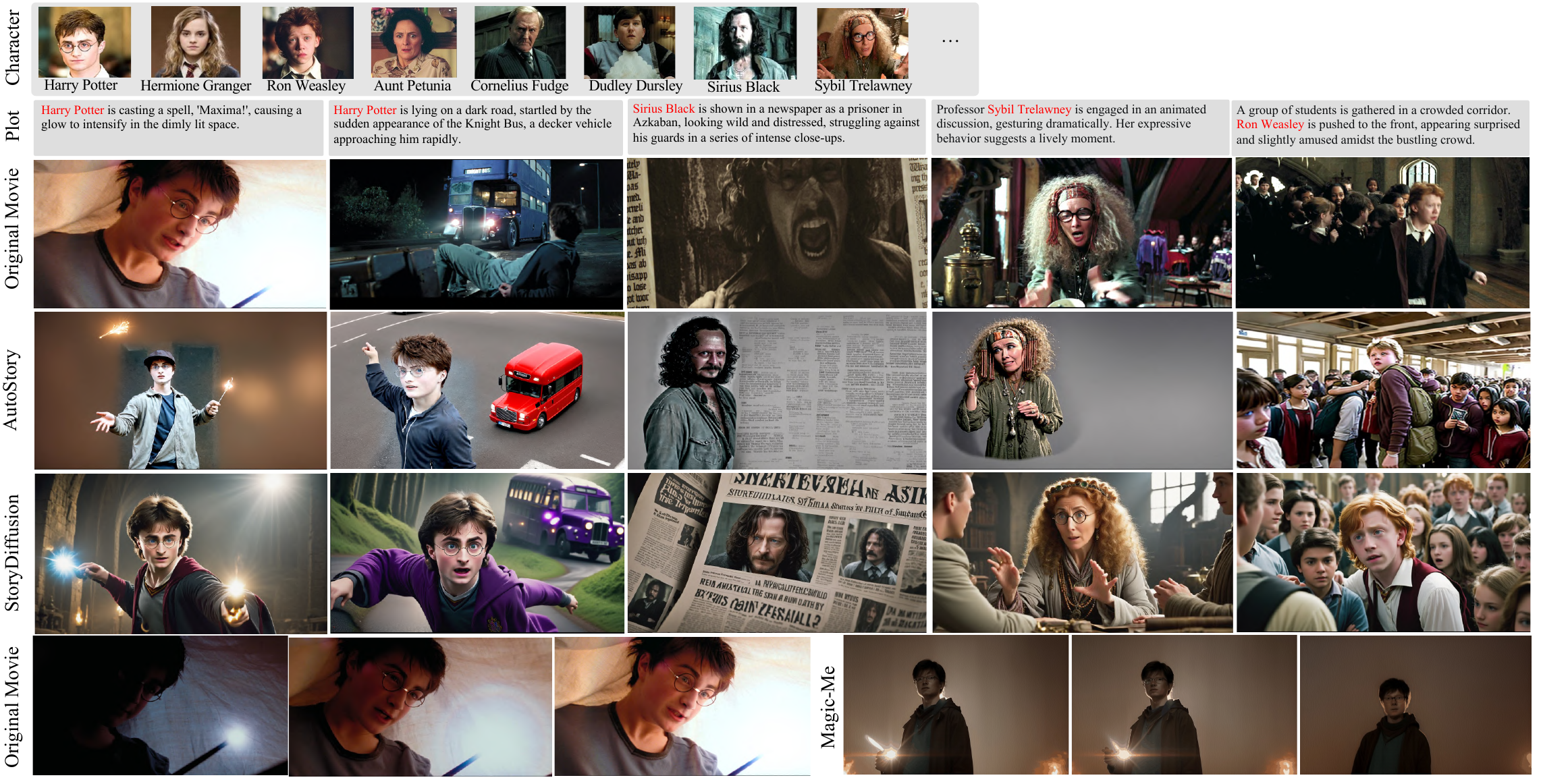}
	\caption{\textbf{Visualization for Text to Keyframe/Video Generation.} 
     \Ours enables a reconstruction of the movie plot.
    }
\label{Visualization}
\end{figure*}

\subsection{Metrics}
\label{Metric}

%
Commonrics include: CLIP Score~\cite{taited2023CLIPScore} measures the correlation between video and text description;
Aesthetic Score, which quantifies the overall visual appeal; 
Frechet Image Distance (FID)~\cite{Seitzer2020FID}, assesses the frame realism by comparing distribution;
and the Inception Score.
%
However, metrics like CLIP Score, can not assess the coherent storyline or character consistency.
Therefore, we introduce three metrics: ${\rm  Precision_{ID} }$,  ${\rm  Recall_{ID} }$, and ${\rm  F1\,\,Score_{ID} }$. 
The metrics evaluate the consistency of character in the generated videos by comparing them to the character list of script.
%
Specifically, we first use DeepFace~\cite{serengil2020lightface} to detect and recognize characters in the generated video, obtaining a set of characters $\sum_{i=1}^{n} \mathcal{C}^{\text{pred}}_i$, where $n$ and $i$ refer to the total number and $i$-th video shot.
We then compute precision, recall, and F score by comparing $\sum_{i=1}^{n} \mathcal{C}^{\text{pred}}_i$ with the set of characters $\sum_{i=1}^{n} \mathcal{C}^{\text{gt}}_i$.
With the sets $\sum_{i=1}^{n} \mathcal{C}^{\text{gt}}_i$ and $\sum_{i=1}^{n} \mathcal{C}^{\text{pred}}_i$, we can calculate the following metrics: 
the False Positives (FP) as ${\rm  FP} : \sum_{i=1}^{n} \left| \mathcal{C}^{\text{pred}}_i \setminus \mathcal{C}^{\text{gt}}_i \right|$, 
the False Negatives (FN)
as ${\rm  FN} : \sum_{i=1}^{n} \left| \mathcal{C}^{\text{gt}}_i \setminus \mathcal{C}^{\text{pred}}_i \right|$, 
and the True Positives (TP). 
%
Finally, ${\rm Recall_{ID}} : \frac{{\rm TP}}{{\rm TP} + {\rm FN}}$, ${\rm Precision_{ID}} = \frac{{\rm TP}}{{\rm TP} + {\rm FP}}$ 
and ${\rm F1_{ID}} : \frac{{\rm  Recall_{ID}} \times {\rm  Precision_{ID}} \times 2}{{\rm  Recall_{ID}} + {\rm  Precision_{ID}}}$ can be calculated.

%% file: video_quality.tex
\begin{tabular}{c|ccc}
Dataset  & Aesthetic Score  $\uparrow$ &  Inception Score $\uparrow$ & \\
    \hline
   InternVid~\cite{internvid}  & 9.09  &  11.68 \\
   MiraData~\cite{ju2024miradata}  &  9.70 & 6.27  \\
   \Ours  & 20.67 & 12.34  \\

    \end{tabular}

%% file: EvalforShotLevel.tex
\begin{tabular}{l|c|cc}
    \multirow{2}{*}{Description} & \multirow{2}{*}{Model}   & \multicolumn{2}{c}{Performance~(Score 1-5)}  \\
    \cline{3-4} 
    & & Completeness $ \uparrow $ &  Hallucination $ \downarrow $ \\
    \hline
    \multirow{3}{*}{Plot}  & MiniCPMV  & 4.12 & 3.14\\
    &  Gemini-1.5-pro   & 4.58 & 1.76  \\  
     & GPT4-o & \textbf{4.78} & \textbf{1.74}    \\
    \hline
    \multirow{3}{*}{Background}  & MiniCPMV  & 4.36 & 2.52  \\
    &  Gemini-1.5-pro   & 4.64 & \textbf{1.17} \\
     & GPT4-o & \textbf{4.81} & 1.24    \\
    \hline
    \multirow{3}{*}{Camera}  & MiniCPMV  & 4.32 & 2.48  \\
    &  Gemini-1.5-pro   & 4.64 & 1.58\\
     & GPT4-o & \textbf{4.88} & \textbf{1.17}    \\
    \hline
    \multirow{3}{*}{Style}  & MiniCPMV  & 4.60 & 1.61 \\
    &  Gemini-1.5-pro   & 4.88 & \textbf{1.11} \\
     & GPT4-o & \textbf{4.95}  & 1.17   \\
    \end{tabular}

%% file: CharacterShotLevel.tex
\begin{tabular}{c|ccc}
\multirow{2}{*}{Model}   & \multicolumn{3}{c}{${\rm  Character\,\,ID\,\,Consistency /\% \uparrow}$/\%}  \\
    \cline{2-4} 
    & ${\rm  Precision_{ID} }$ &  ${\rm  Recall_{ID} }$ & ${\rm  F1\,\,Score_{ID} }$\\
    \hline
   
     MiniCPMV  & 23 & 66 & 34 \\
     Gemini-1.5-pro   & 90 & 76 & 82\\
      GPT4-o & 90 & 97 & 93  \\

    \end{tabular}

%% file: 4_experiment.tex
\section{Experiments}
\label{Experiments}



\subsection{Text to Keyframe/StoryBoard Generation}

Text-to-Keyframe/Storyboard Generation refers to the creation of coherent visual sequences, where character consistency is maintained.
\textbf{Baseline:}
As shown in Table~\ref{tab:experiment}, StoryGen~\cite{liu2024intelligent},StoryDiffusion~\cite{zhou2024storydiffusion}, and AutoStory~\cite{wang2023autostory}, as the three commonly keyframe generation models are evaluated.
Since the code for StoryDiffusion~\cite{zhou2024storydiffusion} is not publicly available, we test with the official pretrained weights directly on the test set.
For the LoRA-based~\cite{ryu2023low} AutoStory~\cite{wang2023autostory}, we trained a unique LoRA model for each character.
\textbf{Analysis:}
Given the plot of each shot-level video, the model generates a keyframe, as illustrated in Figure~\ref{Visualization}.
For evaluating specific metrics (\eg FID, clip score), we extracted the middle frame from the video as the ground truth.
Table~\ref{tab:experiment} shows that generated keyframes still struggle to maintain character consistency, making it challenging to meet real-world application.
We do not compare with pure text-to-video generation methods such as NUWA-XL~\cite{yin2023nuwa} and Phenaki~\cite{villegas2022phenaki}, as they lack character consistency and are not open-source.

\begin{table*}[t]
    \centering
    \small 
    \setlength{\tabcolsep}{1mm}
    \caption{\textbf{Performance for Text/Image to Keyframe/Video Generation on \Ours.} 
    Models without character consistency~(\eg Open-Sora~\cite{opensora}) are excluded.
    %
    %
    %
    `Sub Cons.', `Bg Cons.', `M Smth.', and `Dyn.' refer to `Subject Consistency', `Background Consistency', `Motion Smoothness', and `Dynamic Degree' from the advanced VBench Metrics~\cite{vbench}, respectively.
    %
    }
    \input{experiment}
    \label{tab:experiment}
\end{table*}


\begin{figure*}[t]
	\includegraphics[width=0.99\linewidth]{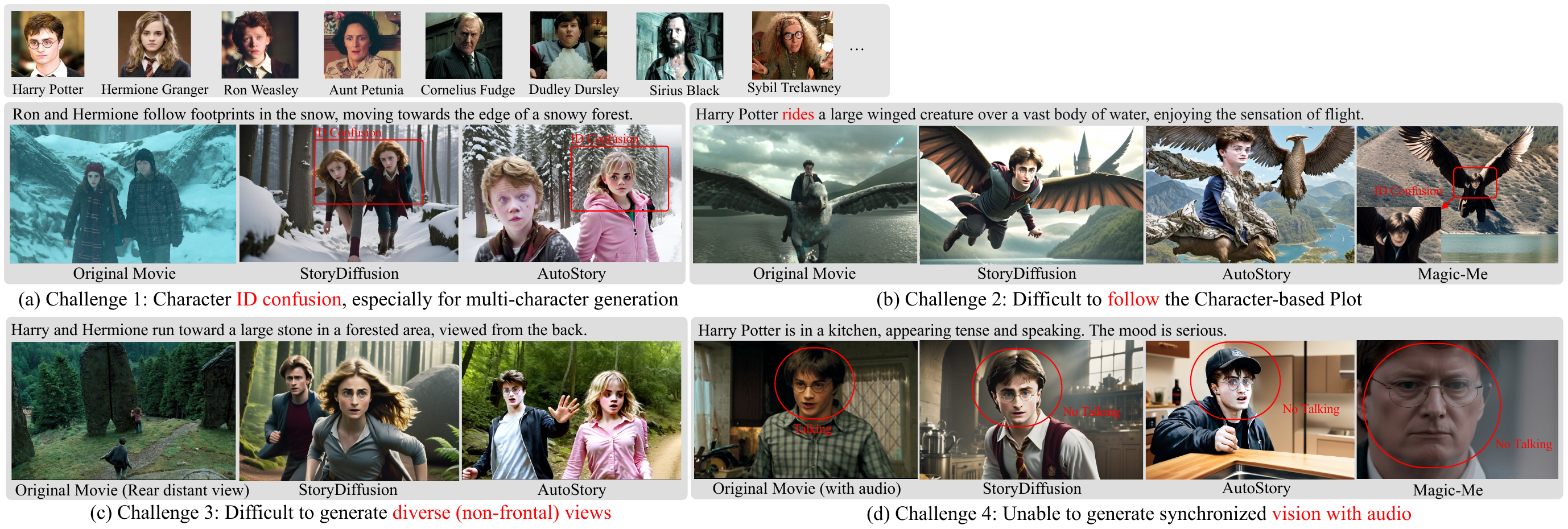}
	\vspace{-0.2cm}
	\caption{\textbf{New Challenges for Movie-Level Keyframe/Video Generation.} 
    }
\label{Challenges}
\end{figure*}

\begin{figure*}[t]
    \centering
    \begin{minipage}{0.63\textwidth}
        \centering
        \small 
    
    \includegraphics[width=\textwidth]{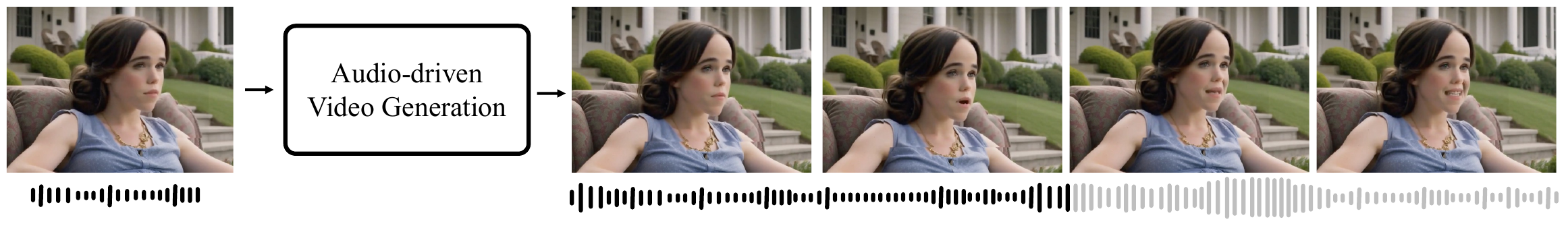} 
    \vspace{-0.7cm}
        \caption{\textbf{Visualization for Audio-driven Talking Human Generation}.}
        \label{fig:talking_head}
    \end{minipage}\hfill
    \begin{minipage}{0.36\textwidth}
        \small
        \centering
        \input{talkinghuman}
        \captionof{table}{\textbf{Talking Human Generation}.}
        \label{tab:talkinghuman}
    \end{minipage}
\end{figure*}

\subsection{Identity-Customized Long Video Generation}
Identity-Customized Long Video Generation involves creating long videos that consistently feature specific characters.
\textbf{Baseline:} DreamVideo~\cite{wei2024dreamvideo} and Magic-Me~\cite{ma2024magic} were used to evaluate our dataset.
During training, only the character bank is used to learn the features of each character.
\textbf{Analysis:} Table~\ref{tab:experiment} demonstrates that the generated characters struggle significantly with consistency.
Several factors contribute to this: 1) Video generation is more challenging than image generation, and current models~\cite{zhang2023i2vgen} often fail to produce coherent videos.
2) Both methods can maintain consistency for only one character per video, limiting their ability to handle multiple characters simultaneously.

\subsection{Image-conditioned Video Generation}
Image-conditioned video generation involves producing subsequent frames based on an initial frame.
\textbf{Baseline:} I2VGen-XL~\cite{zhang2023i2vgen}, SVD~\cite{svd}, and CogVideoX~\cite{yang2024cogvideox} are used as baselines, requiring the first frame of the real video as the conditioning input.
\textbf{Analysis:} Due to the use of real images, the generated videos are closer to real videos, which results in better FID and FVD scores. 
However, the {${\rm  F1\,\,Score_{ID} }$} is unsatisfactory for two main reasons: 
1) The first frame may not include all the characters that appear in the shot. 
2) Movies often feature shots from various angles, as shown in Figure~\ref{Challenges} (c), and the current model cannot effectively identify and maintain consistency of the character.

\subsection{Talking Human with Audio Generation}
Talking Human with Audio Generation refers to the task of creating video scenes featuring a specific character talking, based on provided subtitles or audio.
\textbf{Baseline:} 
Edtalk~\cite{tan2025edtalk} and Hallo2~\cite{cui2024hallo2}, two widely-used audio-driven video generation models, were applied to validate our dataset.
\textbf{Analysis:} Current audio-driven talking generators still primarily focus on generating talking head animations; creating full-body or multi-person talking scenes remains a significant challenge.
As shown in Figure~\ref{tab:talkinghuman}, ensuring audio and visual consistency with synchronized lip movements for multiple characters is highly challenging.

\subsection{Ablation and Analysis}

\textbf{Emerging Challenge in Multi-Characters Consistency.}
%
Table~\ref{tab:experiment} shows that existing models struggle with multi-characters consistency, achieving a maximum accuracy of only $53\%$.
Figure~\ref{Challenges} (a) presents visual comparisons that highlight significant limitations in current models for multi-character generation.
Additionally, StoryDiffusion and Magic-Me maintain consistency for only one character at a time, struggling with multi-character consistency.
Autostory~\cite{wang2023autostory} uses Mix-of-Show~\cite{gu2024mix} with pose-guided techniques for character ID management, but obtaining pose sequences in real-world applications is challenging.



\textbf{Emerging Challenge in Character-based Plot Following.}
\Ours has introduced new challenge in character-based plot following, particularly in defining character relationships and accurately generating interactions of characters. 
Figure~\ref{Challenges} (b) illustrates the difficulties in following these plots and understanding character interactions.
For instance, methods like Autostory, which rely on explicit constraints, often fail to generate distinct human poses and struggle to interpret actions like `ride'.
%

\textbf{Emerging Challenge from Diverse Views Generation}
Figure~\ref{Challenges} (c) highlights another challenge: generating scenes and characters from various views while maintaining consistency in appearance.
Current models can only produce close-up, frontal shots of characters, struggling with other angles or wide shots. 
However, diverse views are essential in movies to convey different moods and purposes.

\textbf{Emerging Challenge in Synchronized Video Generation with Audio.}
No existing method or dataset has explored the generation of synchronized video with audio in long videos,
as shown in Figure~\ref{Challenges} (d).
Some works, such as PersonaTalk~\cite{zhang2024personatalk} and Hallo2~\cite{cui2024hallo2}, have explored and advanced the task of generation of talking heads.
However, these existing tasks focus on single-person talking head generation, leaving a substantial gap when it comes to realistic multi-character dialogue in movie contexts.
%

%% file: experiment.tex
\begin{tabular}{l|ccccc|cccc|ccc}
    \multirow{2}{*}{Method}  & \multirow{2}{*}{CLIP$ \uparrow $} & \multirow{2}{*}{Inception$ \uparrow $}  & \multirow{2}{*}{Aesthetic$ \uparrow $} & \multirow{2}{*}{FID$ \downarrow $ } & \multirow{2}{*}{FVD$ \downarrow $ } & \multicolumn{4}{c|}{VBench Metircs~\cite{huang2024vbench}/\% $\uparrow$ } & \multicolumn{3}{c}{${\rm  Character\,\,Consistency /\% \uparrow}$ }  \\
    \cline{7-13} 
    & & & & &  & Sub Cons. & Bg Cons. & M Smth.& Dyn. & {${\rm  Precision_{ID} }$}  &  {${\rm  Recall_{ID} }$} & {${\rm  F1_{ID} }$}
    \\
    \hline
    \multicolumn{12}{l}{\textit{Text to Keyframe/StoryBoard Generation}}  \\
    \hline
    
    StoryGen &  20.37 &  7.01 &  22.46 & 16.17 & - & - & - & -& -& 77.00 & 1.40 & 2.80  \\
    StoryDiffusion & 21.56 & 9.13 & 26.08 & 11.84 &  - & - & - & -& - & 78.17 & 37.26 & 50.47 \\

    AutoStory &  20.16 & 7.14 & 23.87 & 18.68 & - & - & - & -& - & 77.61 & 41.81 & 54.34 \\
    \hline

    \multicolumn{12}{l}{\textit{Text to Video Generation}}  \\
    \hline
    
    DreamVideo& 22.39  & 11.63 & 19.13  &  7.99 & 853.36& 88.05& 92.97& 96.19  & 69.47 
    & 8.07 & 2.43 & 3.74 \\

    Magic-Me &   21.52 &  10.81 & 20.87  & 8.63 & 789.12 & 96.90 & 96.31 & 98.24& 15.97 & 41.30  & 5.80  & 10.17 \\
    \hline

    \multicolumn{12}{l}{\textit{Image to Video Generation}}  \\
    \hline
    I2VGen-XL &   22.39 & 8.63 & 8.12 & 1.77 & 512.47& 76.18 & 85.38 & 96.30 & 79.16 & 19.41& 10.82 &13.90  \\
    
    SVD &  22.28  & 7.36 &   11.45 & 1.25 & 190.48 & 92.97& 94.48& 98.17 & 84.67 & 20.49 & 12.54 & 15.56 \\
    
    CogVideoX &  22.43  &  7.54 &  14.16 & 1.23  &228.80 & 90.37 & 93.78 & 98.60  & 50.42 & 24.80 & 15.63 & 19.17 \\
\end{tabular}

%% file: talkinghuman.tex
\begin{tabular}{c|ccc}
Model & FID $ \downarrow $ &  FVD $ \downarrow $ & Lip-sync $ \downarrow $\\
    \hline
   
     Edtalk~\cite{tan2025edtalk}  & 2.39 & 504.49 & 2.41 \\
     Hallo2~\cite{cui2024hallo2}   &1.68 & 475.13 & 1.66

    \end{tabular}

%% file: 5_conclusion.tex
\section{Conclusion}
This paper presents \Ours, a hierarchical movie-level dataset specifically designed for long video generation.
%
%
We reevaluated existing generation models and identified new insights and challenges in the field, including the need to maintain consistency across multiple character IDs, ensure multi-view character coherence, and align generated results with complex storylines.
%
%
By facilitating the exploration of complex character interactions and rich storylines, \Ours aims to advance research in long video generation and inspire future developments in the field.

%% file: X_suppl.tex
\clearpage
\setcounter{page}{1}
\maketitlesupplementary

\section{User Study}
\begin{figure}[t]
	\includegraphics[width=0.99\linewidth]{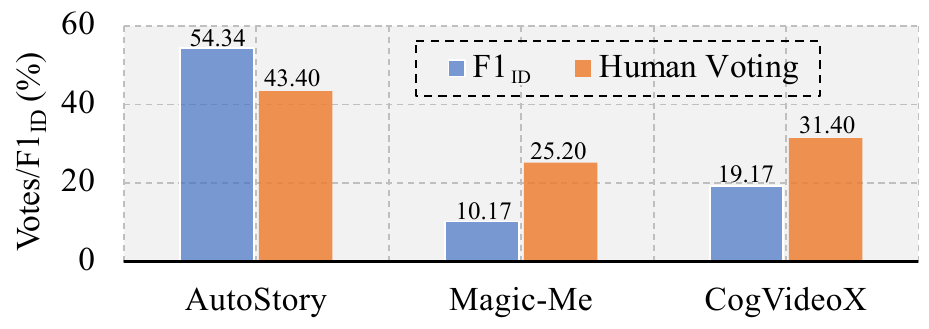}
	\caption{\textbf{User Study $v.s.$ Automatic Metric.} 
     Automatic metrics and human evaluations show a positive correlation.
    }
\label{ScenesDistribution11}
\end{figure}

To examine the correlation between automatic evaluation metrics and human judgment, we compare {${\rm  F1_{ID} }$} scores with human voting results across three models: AutoStory, Magic-Me, and CogVideoX, as shown in Figure~\ref{ScenesDistribution11}.
The results indicate a positive correlation between the two evaluation methods, with models achieving higher {${\rm  F1_{ID} }$} scores also receiving more favorable human votes.
Specifically, AutoStory demonstrates the highest alignment, with an {${\rm  F1_{ID} }$} of 54.34\% and a human voting score of 43.40\%, suggesting strong consistency between automatic and human evaluations.
While Magic-Me and CogVideoX also follow this trend, their human evaluation scores exceed their {${\rm  F1_{ID} }$} values by a notable margin, indicating that certain qualitative factors influencing human preference may not be fully captured by automatic metrics. 
%

\section{Customized Audio Generation}
Customized Audio Generation involves creating customized soundtracks for specific characters and emotional cues.
We conduct experiments on $6$ movies from the test set, splitting the audio of each character into two parts: half as the test set and half as reference audio for evaluation.
Following prior works~\cite{chen2024f5,casanova2024xtts}, we evaluate performance on a cross-sentence task, where the model synthesizes a reading of a reference text in the style of a given speech prompt.

\begin{table}[t]
    \centering
    \small 
    \setlength{\tabcolsep}{1mm}
    \caption{\textbf{Performance for Customized Audio Generation on \Ours.}}
    \input{audiogen}
    \label{tab:audiogen}
\end{table}

\subsection{Metric}
Following prior work, three common metrics, namely Word Error Rate (WER), Speaker Similarity (SIM-o), and Mel Cepstral Distortion (MCD), are used to evaluate our dataset.
For WER, Whisper-large-v3~\cite{radford2023robust} is used to transcribe the audio to text, after which word error is calculated at the text level.
For SIM-o, a WavLM-large-based speaker verification model~\cite{chen2022large} is used to extract speaker embeddings, enabling cosine similarity calculation between synthesized and ground truth speech.
For MCD, an open-source PyTorch implementation~\footnote{https://github.com/chenqi008/pymcd} is used to evaluate the similarity between synthesized and real audio.
For evaluation, each audio file is converted to a single-channel, 16-bit PCM WAV with a sample rate of 22050 Hz.

\begin{table}[t]
    \centering
    \small 
    \setlength{\tabcolsep}{1mm}
    \caption{\textbf{Quality Evaluation for Portrait Image of Character on Movie Level.} Character bank demonstrates excellent performance in both portrait quality and name relevance. }
    \input{CharacterPhoto}
    \label{tab:CharacterPhoto}
\end{table}

\begin{figure*}[t]
	\centering
 \includegraphics[width=0.80\linewidth]{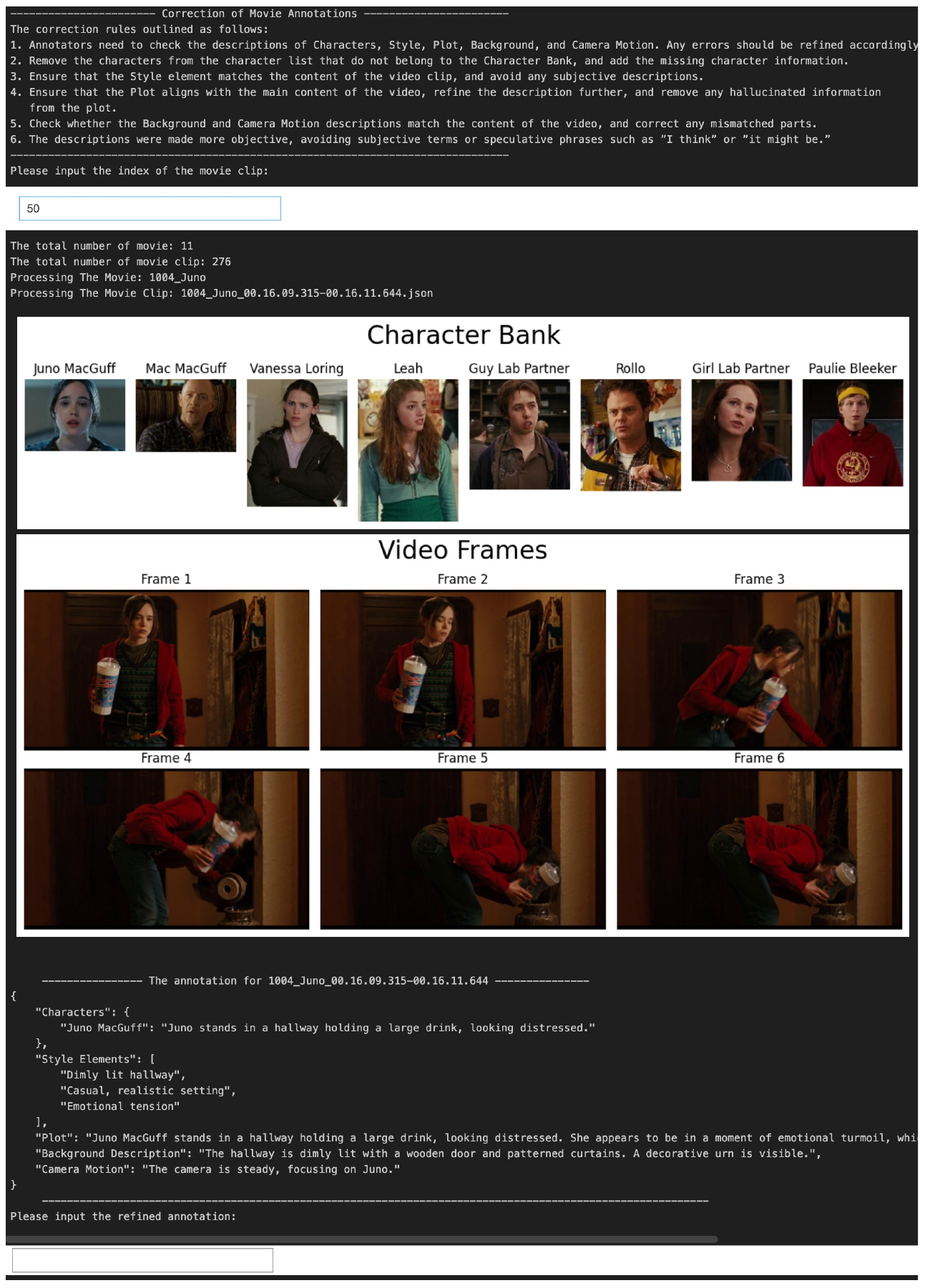}
	\vspace{-0.2cm}
	\caption{\textbf{Manual Correction for Shot-Level Movie Annotations.} 
    }
    \vspace{-0.25cm}
\label{ManualCorrection}
\end{figure*}

\begin{figure*}[t]
	\includegraphics[width=0.99\linewidth]{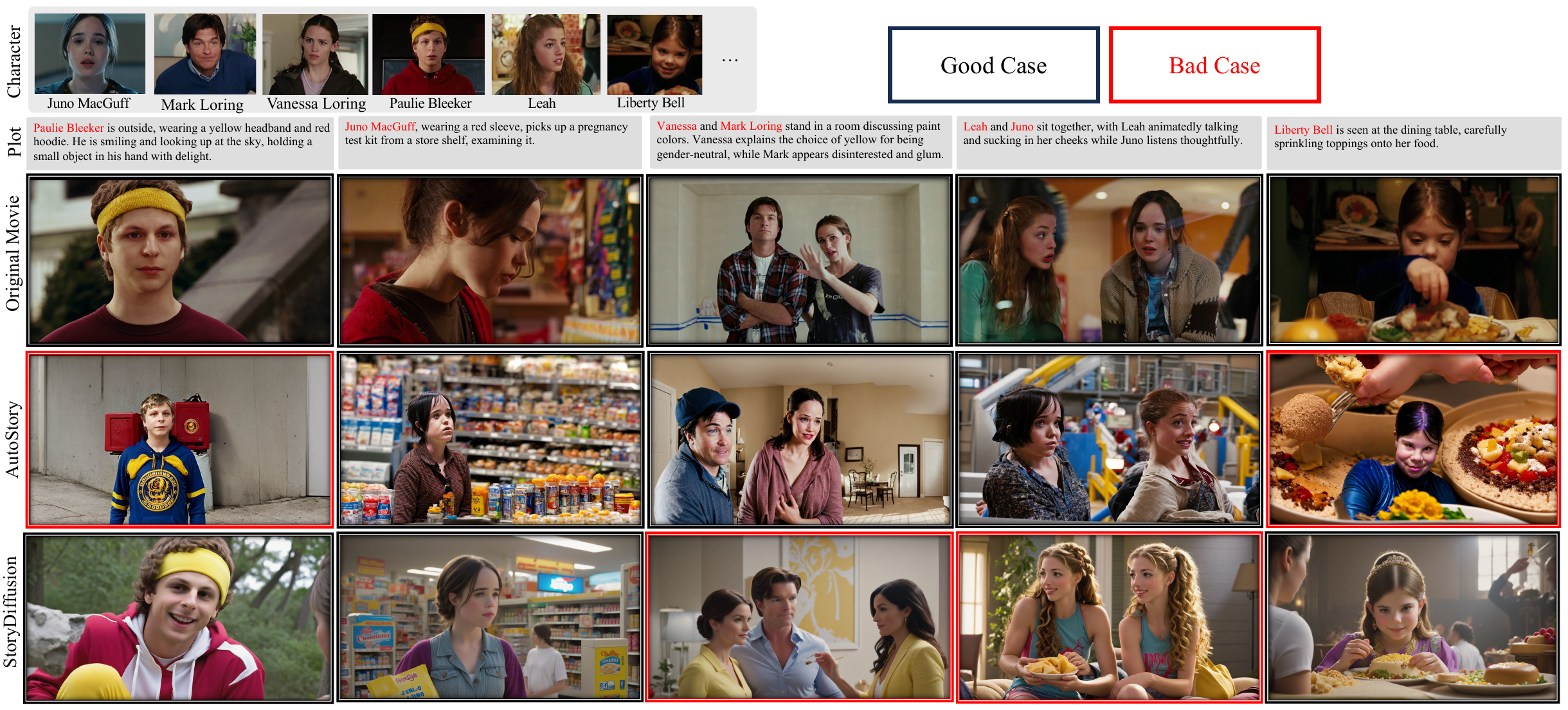}
	\vspace{-0.2cm}
	\caption{\textbf{Visualization Comparison for Movie-Level Keyframe Generation.} 
    }
\label{vis_comparison_1}
\end{figure*}

\begin{figure}[t]
	\includegraphics[width=0.99\linewidth]{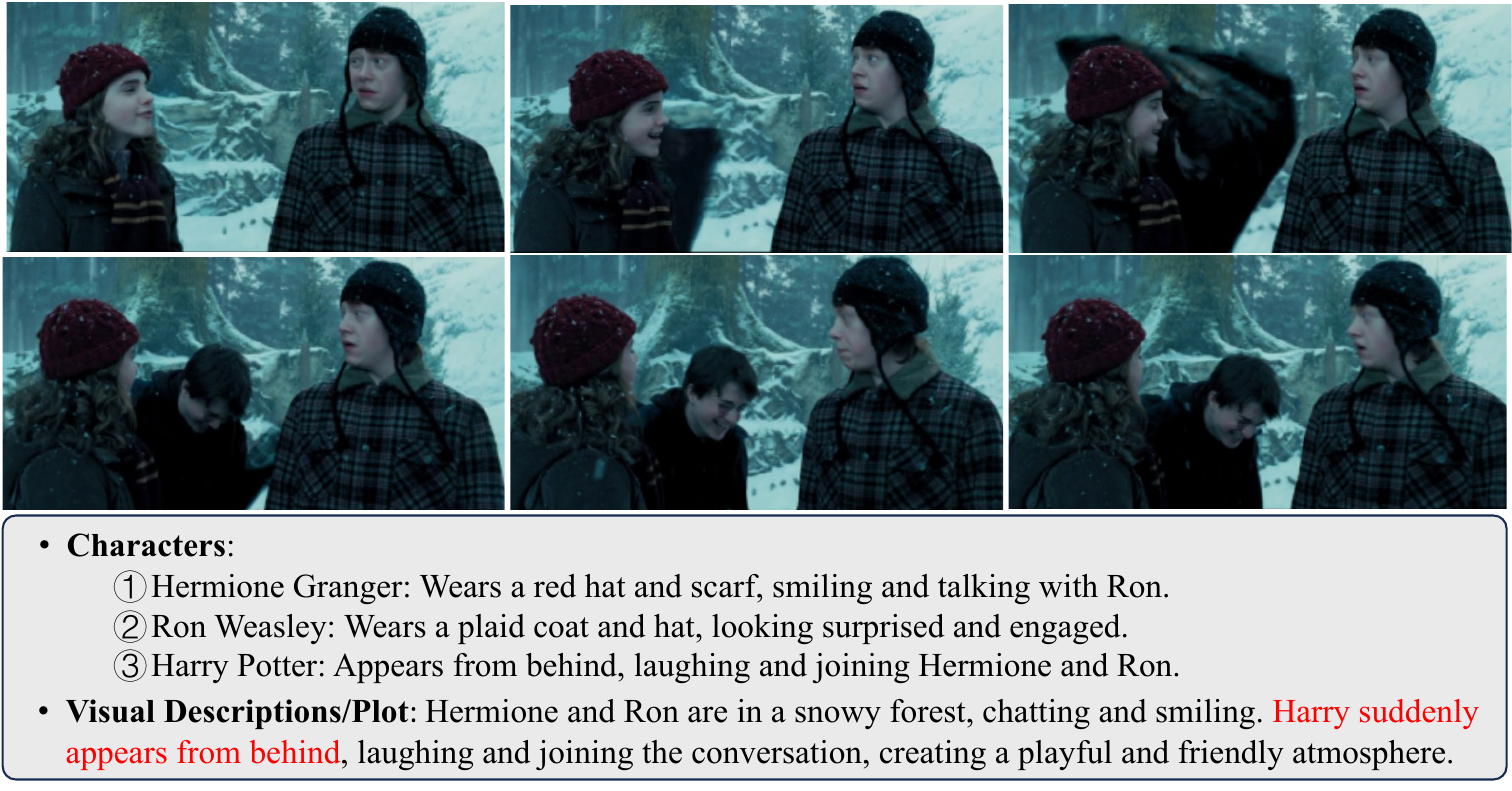}
	\vspace{-0.2cm}
	\caption{\textbf{Temporal Plot Description.} 
     Shot-level plot descriptions often contain strong temporal information that may not be easily represented by a single key frame.
    }
\label{oneortwo}
\end{figure}

\subsection{Baseline and Analysis}
The four audio customization methods—YourTTS~\cite{casanova2022yourtts}, xTTS~\cite{casanova2024xtts}, VALL-E-X~\cite{zhang2023speak}, and F5-TTS~\cite{chen2024f5} were used in MovieBench for evaluation.
We performed direct zero-shot testing without any additional training, with F5-TTS achieving the best performance, as shown in Table~\ref{tab:audiogen}.
Notably, each evaluation was conducted individually for each character.
However, the real challenge lies in scenes with multi-character interactions, as seen in movies.
Generating audio that matches the tone and voice of each character in a way that ensures consistency with the visuals presents a significant challenge, especially in maintaining distinct voices across different audio tracks.

\section{Quality Evaluation and Correction for Shot-Level Annotation}

\textbf{Correction for Description-based Annotations.}
The main text mentions that we required two annotators to manually correct the shot-level dataset in the test set. The specific correction rules are as follows:
\begin{itemize}
    \item \textbf{Check and Refine Descriptions}: reviewing the descriptions of characters, style, plot, background, and camera motion, correcting any inaccuracies.
    
    \item \textbf{Character Set Adjustments}: Characters not belonging to the Character Bank were removed from the video clip’s character set, and any missing characters were added.
    
    \item \textbf{Style Matching}: Ensure that the style element accurately reflected the video clip's content, avoiding subjective interpretations.

    \item \textbf{Plot Alignment}: The Plot was verified to align with the main content of video, with any hallucinated or irrelevant information removed.

    \item \textbf{Grammatical Accuracy}: Descriptions were refined to ensure grammatical correctness.

    \item \textbf{Objectivity}: The descriptions were made more objective, avoiding subjective terms or speculative phrases such as "I think" or "it might be."
\end{itemize}
Two annotators were instructed to progressively refine the character set, style, and plot based on the above rules.
The refinement interface is shown in Figure~\ref{ManualCorrection}.
The interface provides character photos, names, key frames from the original video, and shot-level annotation details (such as plot, appearing character set, etc.).
Annotators use this information to assess the accuracy of the annotations and identify areas for improvement.
The refinement process took two annotators approximately one week.

\begin{figure}[t]
	\includegraphics[width=0.99\linewidth]{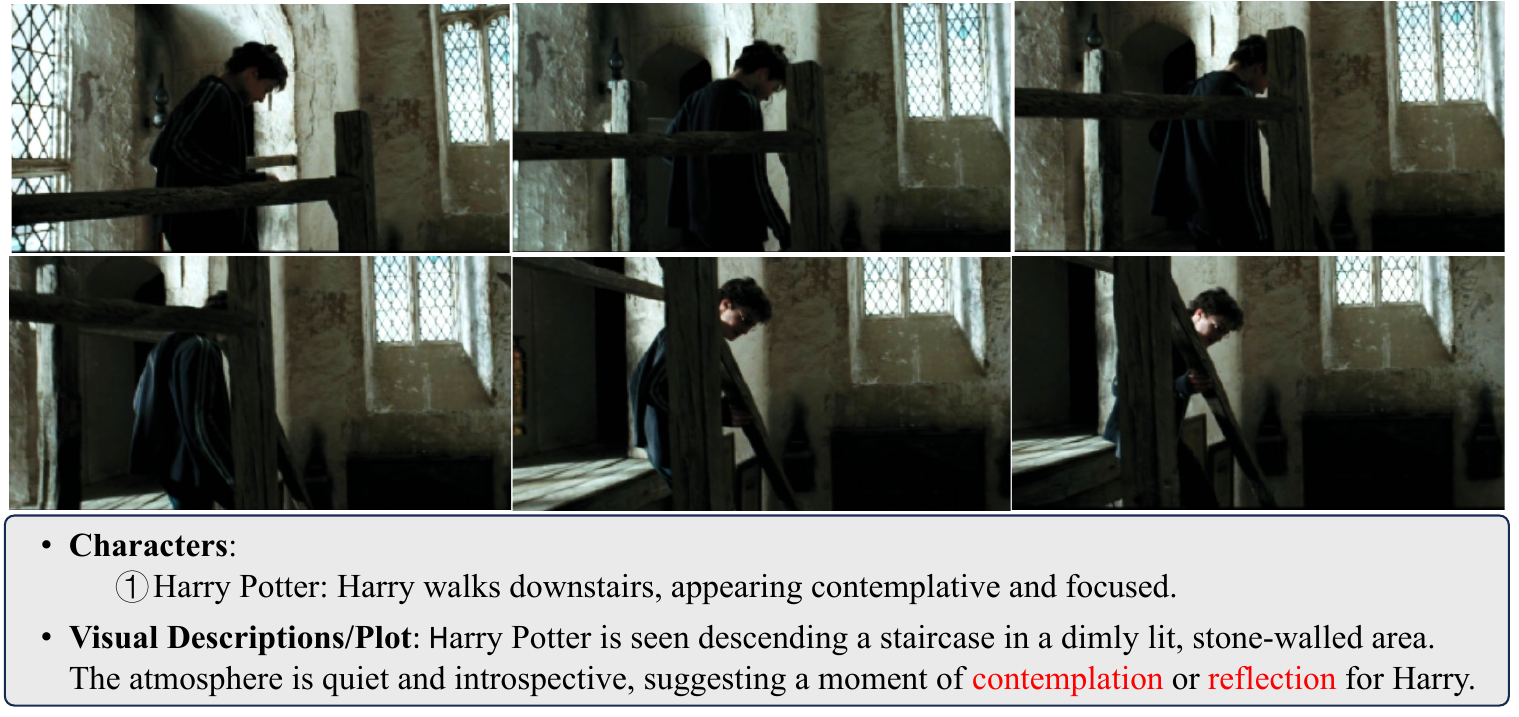}
	\vspace{-0.2cm}
	\caption{\textbf{Hallucination of Generated Plot.} 
     Descriptions generated by GPT-4-o may still exhibit instances of hallucination.
    }
\label{Hallucination}
\end{figure}

\begin{table*}[t]
    \centering
    \small 
    \setlength{\tabcolsep}{1mm}
    \caption{\textbf{Metrics for  Evaluation of Model/Dataset.} `Portrait Quality', `Portrait-Name Relevance', `Completeness', and `Hallucination' are used to assess the quality of MovieBench annotations. 
    Other metrics are primarily used to evaluate model performance.
    }
    \input{Metric_Task}
    \label{tab:Metric_Task}
\end{table*}

\textbf{Quality Evaluation for Shot-level Appearing Character Set.} 
The main text presents that the character photos in our Character Bank were manually selected by two annotators.
After completing the data annotation, we conducted a quality assessment focusing on Portrait Quality and Name-Portrait Relevance.
Table~\ref{tab:CharacterPhoto} shows the relevant experimental results.
It can be observed that Portrait-Name Relevance scores significantly higher than Portrait Quality. 
This is mainly because manually selected images are generally consistent with their names, leaving little room for error in relevance.
However, image quality is harder to guarantee, as not all image candidates are of consistently high quality.

\section{Possible Directions? Single-Stage or Two-Stage}
Movie/long video generation is typically not done in one go; instead, it is divided into multiple shot clips for individual generation.
Currently, there are two main approaches for this task: one-stage and two-stage methods.
\label{baseline}

\textbf{One Stage.}
Currently, there are no fully realized one-stage solutions for this task. 
Most open-source one-stage models~\cite{opensora,zhang2023i2vgen} focus on text-to-video generation, lacking the ability to maintain character consistency and connect storylines across different video clips.
DreamVideo~\cite{wei2024dreamvideo} and Magic-Me~\cite{ma2024magic}, two commonly used customizable video generation models, are utilized in our paper.
The typical workflow involves first creating a script with character-specific details for each shot, generating corresponding video clips for each shot individually, and then stitching these clips together to produce a cohesive long-form video.
%
%





\textbf{Two Stages.}
Directly generating long-form videos is highly challenging.
Therefore, the two-stage strategy has become a more practical solution:
1) Firstly, Key frames/Storyboard generation models~\cite{wang2023autostory,zhou2024storydiffusion,zhao2024moviedreamer,wu2023paragraph} can be used to generate the key frame for every shot-level video.
Figure~\ref{vis_comparison_1} provides additional visualizations of both successful and challenging cases for AutoStory~\cite{wang2023autostory} and StoryDiffusion~\cite{zhou2024storydiffusion}.
It can be observed that AutoStory~\cite{wang2023autostory} excels in maintaining consistency across multiple characters but struggles with certain background compatibility.
On the other hand, StoryDiffusion~\cite{zhou2024storydiffusion} performs well in generating natural interactions between characters and backgrounds but has difficulty maintaining consistency across multiple characters.
2) With key frames, image-conditioned video generation models (\eg{} SVD), are employed to expand the key frames into full video clips.
Finally, the various video clips are stitched together to form a coherent sequence. 
While this method addresses some issues of video continuity and narrative progression, it still faces difficulties with maintaining a smooth flow across clips and ensuring consistent character representation throughout the film.
However, for certain shots with strong temporal dependencies, it is challenging to rely solely on keyframes for representation. 
\textcolor{red}{Figure~\ref{oneortwo} shows an example where generating only a single keyframe is clearly insufficient to capture the sequence of Harry’s appearance.}






\section{Metric Formulation}
As shown in Table~\ref{tab:Metric_Task}, we summarize and formulate the evaluation metrics relevant to the tasks discussed in this paper.
`Portrait Quality' and `Portrait-Name Relevance' assess the accuracy of the Character Bank annotations, specifically evaluating the precision of manual image selection and labeling. 
`Completeness' and `Hallucination' measure the accuracy of description-based annotations (\eg{} plot and background descriptions), focusing on the completeness of details and hallucinations from VLM descriptions.
The CLIP Score, Aesthetic Score, Frechet Image Distance, and Inception Score evaluate the quality of generated images/videos and their alignment with text descriptions.
Additionally, this paper introduces new metrics—${\rm Precision_{ID} }$, ${\rm Recall_{ID} }$, and ${\rm F1_{ID} }$—to assess character consistency.
`Subject Consistency', `Background Consistency', `Motion Smoothness', and `Dynamic Degree' are recently proposed metrics from VBen~\cite{huang2024vbench}, aimed at evaluating various aspects of generated video.

\section{Limitation}
\textbf{Hallucination of Plot from GPT4-o.}
Although GPT-4-o demonstrates high accuracy and rarely makes errors, its generated plot descriptions can still present issues, such as hallucination.
Figure~\ref{Hallucination} provides a clear example: in this video, Harry walks downstairs, yet there is no evidence to conclude that Harry is engaged in contemplation or reflection. 
However, the summary of GPT-4-o confidently suggests this, introducing a potential misinterpretation. 
Such hallucinations can reduce data reliability, misleading model training and potentially causing unstable convergence when using this data.

%% file: audiogen.tex
\begin{tabular}{c|ccc}
Dataset  & MCD $\downarrow$ & WER(\%) $\downarrow$  &  SIM-o $\uparrow$  \\
    \hline
   
   YourTTS~\cite{casanova2022yourtts}  & 8.41 &  0.26 & 0.97 \\

   xTTS~\cite{casanova2024xtts}  & 8.31 &  0.28 & 0.98 \\

   VALL-E-X~\cite{zhang2023speak}  & 4.48 &  1.25 & 0.97 \\
   F5-TTS~\cite{chen2024f5}  & 3.12 &  0.20 & 0.98 
    \end{tabular}

%% file: CharacterPhoto.tex
\begin{tabular}{l|cc}
    Movie  & Portrait Quality & Name Relevance   \\
    \hline
    AS Good As It Gets & 4.56 & 5.00   \\
    Clerks & 4.20 & 4.92   \\
    Halloween & 4.00 & 4.89   \\
    The Hustler & 4.80 & 4.98   \\
    Chasing Amy & 4.42 & 4.78   \\
    The Help & 4.30 & 5.00   \\
    No Reservations & 4.86 & 4.93   \\
    An Education & 4.70 & 4.85   \\
    \parbox[t]{1.4in}{Harry Potter and the \\ Chamber of Secrets} & 4.73 & 5.00   \\
    Seven Pounds & 4.71 & 4.87   \\
\end{tabular}

%% file: Metric_Task.tex
\begin{tabular}{c|c|p{1.43\columnwidth}}
	Metric   & Better & Description   
    \cr
	\hline

   ${\rm  Portrait\,\,  Quality}$
   & higher  &  Quality assessment for character portraits, involving human raters scoring the image quality on a scale of 1 to 5, with 5 being the highest score. \cr \hline

    ${\rm  Portrait\mbox{-}Name\,\,Relevance}$
   & higher  &  Portrait-Name relevance score for each character name and portrait pair, with human raters on a scale of 1 to 5, and 5 being the highest score. \cr \hline
   
   ${\rm  Completeness}$
   & higher  &  Descriptive completeness score, assessing the extent to which the annotation of textual descriptions (\eg{} Plot, Background Description) reflects the completeness of the video content. \cr \hline

   ${\rm  Hallucination}$
   & lower  &  Fantasy score, assessing the degree of hallucination in textual descriptions of video content (e.g., Plot, Background Description). \cr \hline
   
   ${\rm  CLIP\,\,  Score}$
   & higher  &  The evaluation of semantic alignment between the plot and generated outputs \cr \hline
 
    ${\rm  Aesthetic\,\,Score (AS)}$
   & higher  &  The evaluation for aesthetic quality of an image by extracting visual features using the CLIP and comparing them with a pre-trained aesthetic model to quantify the score. \cr \hline
   
   ${\rm  Frechet\,\,Image\,\,Distance (FID)}$
   & lower  &  The evaluation for the quality of generated images by comparing the feature distribution of features between real and generated images. \cr \hline

   ${\rm  Inception\,\,Score\,\,(IS)}$
   & higher  &  The evaluation for the quality and diversity of generated images by Inception network. \cr \hline

    ${\rm  False\,\,  Postive(FP)}$
   & lower  &  The total number of false positives. Formula:  ${\rm  FP} = \sum_{i=1}^{n} \left| \mathcal{C}^{\text{pred}}_i \setminus \mathcal{C}^{\text{gt}}_i \right|$. 
   \cr \hline
   
    ${\rm  False\,\,  Negative(FN)}$
   & lower  & The total number of false negative. Formula:  ${\rm  FN} = \sum_{i=1}^{n} \left| \mathcal{C}^{\text{gt}}_i \setminus \mathcal{C}^{\text{pred}}_i \right|$.\cr \hline
   
   ${\rm  True\,\, Postive(TP)}$
   & higher  &  The total number of true positives. Formula: ${\rm  TP} = \sum_{i=1}^{n} \left| \mathcal{C}^{\text{gt}}_i \cap \mathcal{C}^{\text{pred}}_i \right|$. \cr \hline

   {${\rm  Recall_{ID} }$}
   & higher
   & Ratio of correct detections\&recognitions to total number of GTs. Formula: $\frac{{\rm  TP}}{{\rm  TP} + {\rm  FN}}$ \cr \hline

   {${\rm  Precision_{ID} }$}
   & higher 
    & Ratio of correct detections\&recognitions to total number of predicted detections\&recognitions. Formula: $\frac{{\rm  TP}}{{\rm  TP} + {\rm  FP}}$ \cr\hline

     {${\rm  F1\,\, Score_{ID} }$} & higher&{${\rm  F1\,\, Score_{ID} }$}~\cite{ristani2016performance}. The ratio of correctly identified detections\&recognitions over the average number of ground-truth and computed detections\&recognitions. Formula: $\frac{{\rm  Recall_{ID}} \times {\rm  Precision_{ID}} \times 2}{{\rm  Recall_{ID}} + {\rm  Precision_{ID}}}$ \cr\hline

   Subject Consistency
   & higher 
    & DINO~\cite{caron2021emerging} is used to assess whether the appearance remains consistent throughout the entire video.\cr\hline

    Background Consistency
   & higher 
    & CLIP feature similarity~\cite{radford2021learning} is used to evaluate the temporal consistency of the background scenes.\cr\hline

    Motion Smoothness
   & higher 
    & Video frame interpolation
    model~\cite{li2023amt} is used to evaluate the smoothness of generated motions.\cr\hline

    Dynamic Degree& higher 
    & Optical flow estimation~\cite{teed2020raft} is used to evaluate the degree of dynamics in synthesized videos.\cr
   

\end{tabular}